%% file: root.tex
\documentclass[letterpaper, 10 pt, conference]{ieeeconf}  % Comment this line out if you need a4paper

\IEEEoverridecommandlockouts                              % This command is only needed if 
\usepackage{times} % assumes new font selection scheme installed
\usepackage{amsmath} % assumes amsmath package installed
\usepackage{amssymb}  % assumes amsmath package installed
\usepackage{mathtools}
\usepackage{graphicx}
\usepackage{hyperref}
\usepackage{setspace}

\input{math_notations}

\newcommand{\customfootnotetext}[2]{{% Group to localize change to footnote
  \renewcommand{\thefootnote}{#1}% Update footnote counter representation
  \footnotetext[0]{#2}}}% Print footnote text

\title{\LARGE \bf
Closing the Perception-Action Loop for Semantically Safe Navigation in Semi-Static Environments
}

\author{\authorblockN{Jingxing Qian\textsuperscript{1},
Siqi Zhou\textsuperscript{2},
Nicholas Jianrui Ren\textsuperscript{1}, 
Veronica Chatrath\textsuperscript{2}, 
Angela P. Schoellig\textsuperscript{1,2}
}
}

\usepackage{amsmath}

\newcommand{\defeq}{\vcentcolon=}

\begin{document}

\maketitle

\customfootnotetext{1}{The authors are with the University of Toronto Institute for Aerospace Studies and the University of Toronto Robotics Institute. \\Email: {\tt\footnotesize \{firstname.lastname\}@robotics.utias.utoronto.ca}}
\customfootnotetext{2}{The authors are with the Technical University of Munich and the Munich Institute of Robotics and Machine Intelligence (MIRMI). \\Emails:
{\tt\footnotesize \{firstname, lastname\}@tum.de}}

\thispagestyle{empty}
\pagestyle{empty}

\begin{abstract}
%\begin{spacing}{0.98}
Autonomous robots navigating in changing environments demand adaptive navigation strategies for safe long-term operation. While many modern control paradigms offer theoretical guarantees, they often assume known extrinsic safety constraints, overlooking challenges when deployed in real-world environments where objects can appear, disappear, and shift over time. In this paper, we present a closed-loop perception-action pipeline that bridges this gap. Our system encodes an online-constructed dense map, along with object-level semantic and consistency estimates into a control barrier function (CBF) to regulate safe regions in the scene. A model predictive controller (MPC) leverages the CBF-based safety constraints to adapt its navigation behaviour, which is particularly crucial when potential scene changes occur. We test the system in simulations and real-world experiments to demonstrate the impact of semantic information and scene change handling on robot behavior, validating the practicality of our approach. %\customfootnotetext{4}{%This work was supported by the Vector Institute for Artificial Intelligence in Toronto and the NSERC Canadian Robotics Network (NCRN).}
%\end{spacing}
\let\thefootnote\relax\footnotetext{This work was supported by the Vector Institute for Artificial Intelligence in Toronto and the NSERC Canadian Robotics Network (NCRN).}
\let\thefootnote\relax\footnotetext{A video description of our framework with additional experiments is available at \url{http://tiny.cc/obj-mpc-cbf.}}
\end{abstract}

\begin{spacing}{1.0}

\section{INTRODUCTION}
\input{sections/introduction}

\vspace{-0.1\baselineskip}
\section{RELATED WORKS}
\input{sections/litreview}

\input{sections/problem_formulation.tex}

\input{sections/methodology.tex}

\input{sections/exp_results.tex}

\section{CONCLUSION}
\input{sections/conclusion.tex}

\end{spacing}
\clearpage
\newpage

\bibliographystyle{IEEEtran}
\bibliography{references.bib}

\end{document}

%% file: math_notations.tex
\newcommand{\set}{\mathcal}
\newcommand{\R}{\mathbb{R}}
\newcommand{\Z}{\mathbb{Z}}

\newcommand{\bx}{\mathbf{x}}
\newcommand{\bu}{\mathbf{u}}
\newcommand{\bA}{\mathbf{A}}
\newcommand{\bB}{\mathbf{B}}
\newcommand{\bC}{\mathbf{C}}
\newcommand{\bD}{\mathbf{D}}

%% file: sections/introduction.tex
\begin{spacing}{0.97}
Autonomous robots face increasing demands for long-term navigation in challenging environments like warehouses, office spaces and public roads, where prior knowledge is limited. Accurate mapping and localization are critical for safe and efficient operation in such scenarios. However, most localization and mapping systems assume a static world, which is rarely the case in reality~\cite{Cadena2016PastPA}. In these environments, robots may encounter both highly dynamic objects, like humans and other robots, and semi-static objects, such as furniture and pallets, which can appear, disappear and shift over time. Semi-static changes are especially hard to detect, and their mishandling can lead to corrupted maps and lost localization, resulting in catastrophic failure such as obstacle collisions. The growing demand for long-term autonomy necessitates robust and adaptive navigation strategies. 

Recent efforts have addressed object-aware localization and mapping in semi-static scenes~\cite{panoptictsdf,QianChatrathPOCD,Fu2023NeuSENS,QianPOVSLAM}. These approaches all estimate a 0-to-1 consistency score for each mapped object based on sensor data, updating the changed regions and objects when necessary. However, these works solely focus on mapping and localization accuracy, leaving a notable gap in leveraging the resulting object-level semantic and geometric information for downstream decision-making tasks to ensure safe navigation.

% In SLAM, semantic labels are often readily available; however, such information is often not accounted for in closed-loop system. In practical scenarios, we often would like robot to achieve certain behaviours 
% In this work, building on our previous work, we aim to close the perception-action loop for semantically

% In this work, we aim to enable robots to safely navigate in an unknown environment in a semantically safe manner. Here, semantics include both the semantic label and the stationarity score of the objects.

An increasing amount of work has been proposed to address the challenge of ensuring safe decision-making for robots operating in changing and uncertain environments~\cite{brunke2022safe}. Examples include learning-based model predictive control~(MPC)~\cite{rosolia2017learning,ostafew2016learning}, learning-enhanced adaptive control~\cite{gahlawat2020l1,chowdhary2014bayesian}, as well as control barrier function (CBF)-based safety filter designs~\cite{ames2019control, wang2018safe}. While these control approaches offer promising theoretical guarantees, a notable limitation lies in their common assumption that safety constraints are known ahead of time~\cite{brunke2022safe}. Many practical robot systems rely on onboard sensors to perceive their surroundings, while extrinsic constraints are inferred in real-time. Though there exist a few perception-based safe learning frameworks~(e.g.,~\cite{xiao2023barriernet, grandia2023perceptive}), they do not yet exploit object-level semantic and geometric understanding of the environment that can be readily distilled from perception and mapping systems. In this work, we aim to bridge this gap and facilitate the practical application of theoretical advancements in safe control within real-world contexts. Our emphasis lies in leveraging object-level semantic and geometric constraints to enhance the adaptability and safety of robotic systems operating in semi-static environments.
\begin{figure}[t]
\centering
  \includegraphics[width=\linewidth]{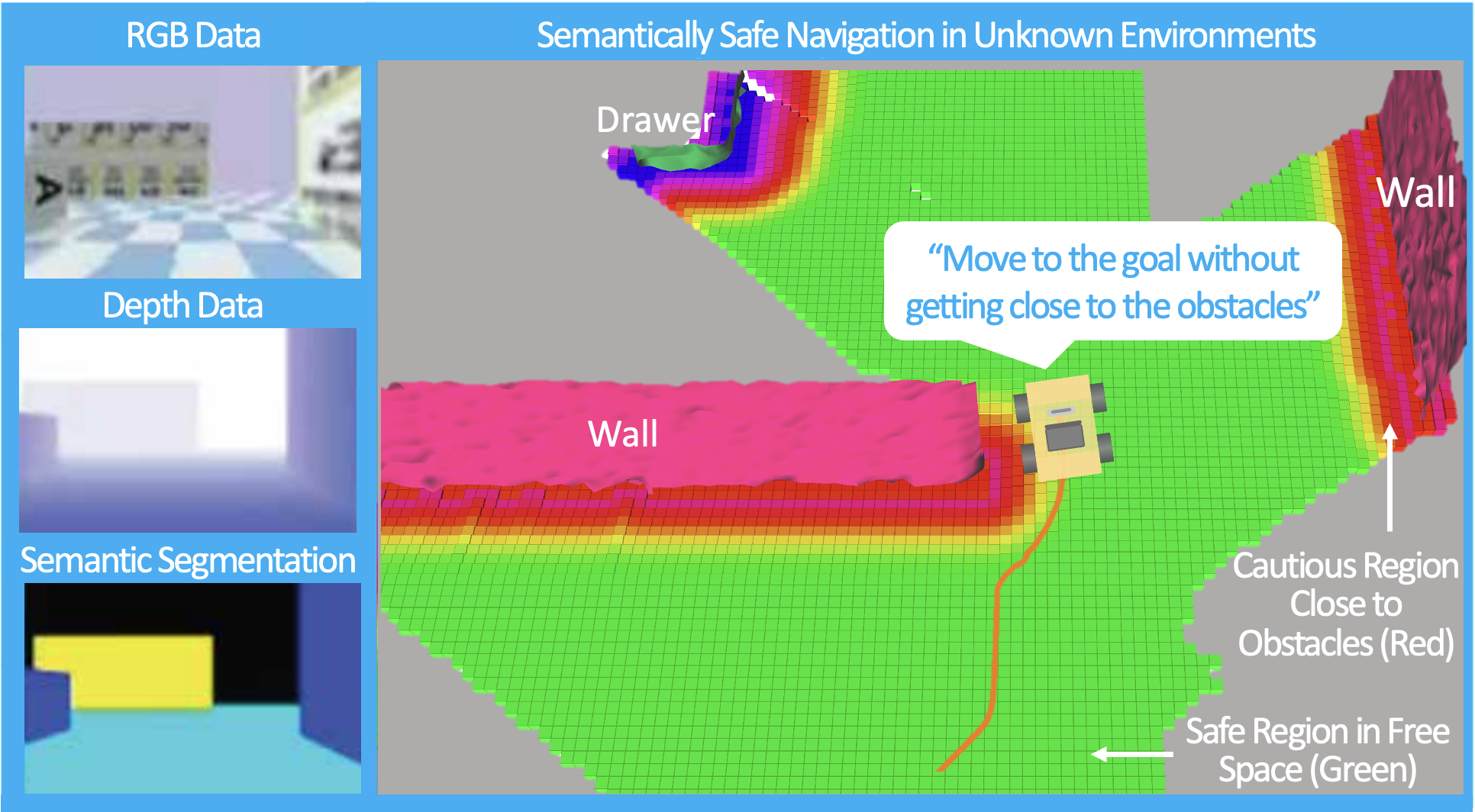}
\caption{Our system takes in semantically annotated RGB-D frames to localize and construct an object-aware volumetric map. Each object holds a semantic label and a consistency score. The map is transformed into a CBF, which is utilized by an MPC to plan safe paths around objects. In our framework, objects that are \textit{likely-static}, such as walls, and those having higher consistency scores, will have smaller unsafe regions around them.}
\vspace{-1.5\baselineskip}
\label{fig:cover}
\end{figure}

We present a closed-loop pipeline that establishes an integrated perception-action loop. Our system adopts the state-of-the-art localization and semi-static mapping strategies~\cite{QianChatrathPOCD,ORBSLAM3_TRO} to construct an up-to-date, object-aware volumetric map (using truncated signed distance functions, TSDFs) of the environment on the fly. Each object within the map maintains a semantic label and a consistency estimate. Upon each update cycle, the map is distilled into a CBF, which synthesizes both the object semantic labels and consistency estimates to regulate safe regions within the scene. Our system further incorporates an MPC that utilizes the CBF-based safety constraints to adapt its navigation strategy around objects. This adaptability is particularly critical when potential scene changes are detected. We conduct tests in both simulated and real-world environments. Through the experiments, we illustrate how using semantic information and accommodating potential scene changes significantly influence the robot's behavior, showcasing the practicality of our approach. Our key contributions are as follows:
\begin{itemize}
\item We develop a method to encode dense volumetric maps, object semantic labels, and object consistency estimates into a single CBF.
\item We build a closed-loop system combining localization, object-aware mapping, and CBF-based safe control for adaptive robot navigation in semi-static scenes.
\item We demonstrate the system's adaptability in both simulated and real-world environments, showcasing dynamic behavior adjustments based on prior semantic knowledge and potential scene changes.
\end{itemize}

\end{spacing}

%% file: sections/litreview.tex
%Charrow \textit{et al.} propose Information-Theoretic Planning with Trajectory Optimization for Dense 3D Mapping in \cite{InformationTheoreticPW}.

% Han \texit{et al.} \cite{Fiesta}

\begin{figure*}[t]
  \centering
  \includegraphics[width=0.92\linewidth]{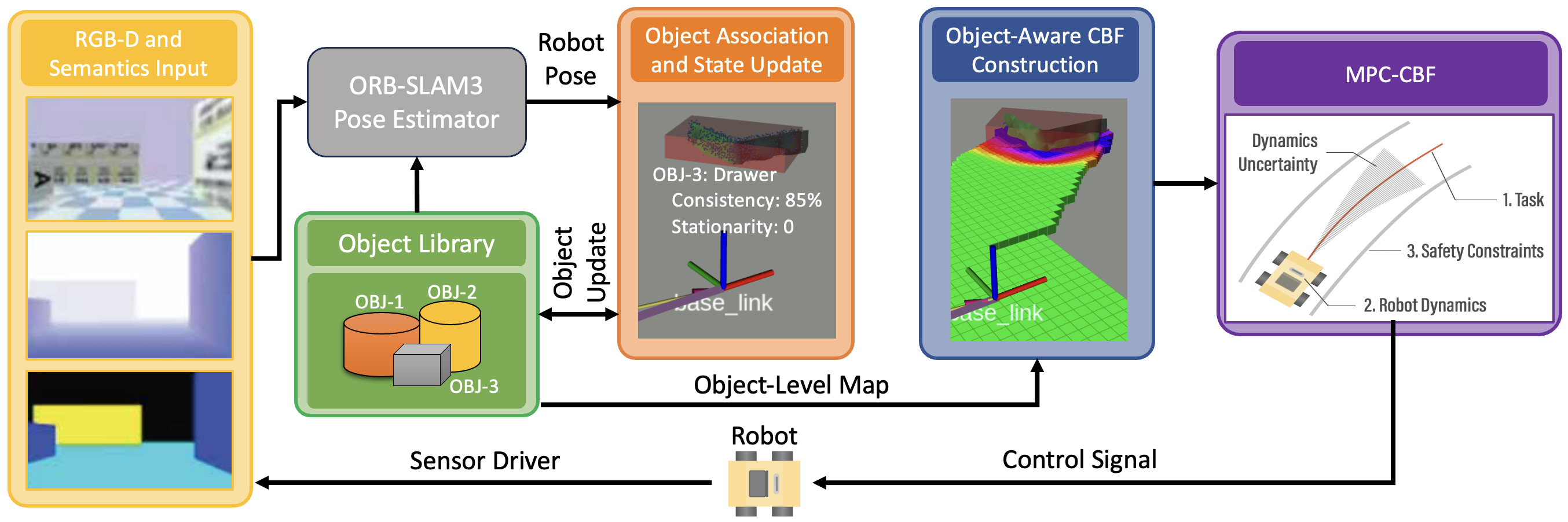}
     \caption{\textbf{Our closed-loop pipeline} (Section~\ref{sec:problem_formulation}) for semantically safe navigation in semi-static environments. The system maintains an object library and takes in semantically segmented RGB-D frames at each timestep. A modified ORB-SLAM3 provides pose estimates using features from \textit{likely-static} objects. From the estimated pose, observations are associated to mapped objects and the consistency is updated for each object (Section~\ref{sec:bayesian_update}). From the joint object map, an object-aware CBF is constructed, that takes object semantics and consistency into account (Section~\ref{sec:construct_cbf}). Finally, an MPC leverages the CBF-based safety constraints to compute actions for the robot (Section~\ref{sec:mpc_cbf}).}
  \label{fig:pipeline}
\vspace{-1.1\baselineskip}
\end{figure*}

\begin{spacing}{0.97}
    
\subsection{Scene Representations and Object-Aware Mapping}

Many geometric representations exist for scene reconstruction, including sparse features from vision systems~\cite{ORBSLAM3_TRO}, surfels~\cite{Pfister2000SurfelsSE}, triangular meshes~\cite{Bloesch2019LearningMF}, point clouds~\cite{dense_vslam_rgbd}, and dense volumetric methods such as occupancy grids~\cite{octomap}, and Signed Distance Functions (SDFs)~\cite{voxblox}. SDFs encode the distance to the nearest surface, gaining traction as they contain rich geometric information for downstream decision-making tasks. SDFs also exploit parallelism to achieve efficient, real-time map updates~\cite{voxblox}. Nevertheless, these methods focus on capturing precise scene geometry, ignoring semantic and object-level knowledge.

Researchers have attempted to fuse semantic and object-level information into the geometric map. SLAM++~\cite{slam++} is an early method which introduces objects into localization and mapping by leveraging existing CAD models from a predefined object library. Voxblox++~\cite{grinvald2019volumetric}, Kimera~\cite{Rosinol20icra-Kimera}, and Hydra~\cite{hughes2022hydra} build on \cite{voxblox} by fusing semantic labels into the scene-level grid representation. Works like Fusion++~\cite{fusion++}, SemanticFusion~\cite{SemanticFusionD3}, and MaskFusion~\cite{Rnz2018MaskFusionRR} leverage semantic segmentation to decompose the scene into object-centric local maps, achieving precise object reconstruction. However, many approaches still assume a static world, rendering them ineffective during real-world deployment where objects may change over time.

\subsection{Change Detection in Mapping and Localization}

To handle dynamics in the scene, two common strategies have been adopted in prior works. The first approach relies on semantic segmentation to identify potentially moving objects, treating them as outliers during camera tracking and scene reconstruction~\cite{Sun2019MovableObjectAwareVS,dsg, DM-SLAM, DOT, DS_SLAM}. Although this approach is highly efficient, it can fail when many potentially changing objects are present. The second approach jointly tracks the camera and potentially dynamic objects~\cite{Rnz2018MaskFusionRR,Hachiuma2019DetectFusionDA,TSDF++,Xu2019MIDFusionOO}. These methods rely on motion consistency between consecutive frames to detect high-rate changes, thus not effective when long-term changes are encountered. 

Detecting semi-static changes is crucial for safe long-term autonomy, but has been overlooked in literature. Existing methods estimate a geometric consistency score for each feature or object in the scene. Panoptic Multi-TSDF~\cite{panoptictsdf} and VI-MID~\cite{visinsMulti} calculate the object consistency based on the overlap ratio between measurements and mapped objects. NeuSE~\cite{Fu2023NeuSENS}, NDFs~\cite{Fu2022RobustCD}, and 3D-VSG~\cite{Looper20223DVL} build local scene graphs to predict object-level changes. POCD~\cite{QianChatrathPOCD} and Rosen~\textit{et al.}~\cite{rosen2016towards} propose Bayesian filters to estimate the object stationarity. POV-SLAM~\cite{QianPOVSLAM} proposes an expectation-maximization method to jointly estimate the robot state and object consistency, but incurs high computation costs.
\vspace{-0.1cm}
\subsection{Safe Decision-Making in Robotics}
Control theory is a basis for providing safety guarantees based on knowledge of the robot system~\cite{brunke2022safe}. There has been an increasing trend of combining control theoretic frameworks and learning-based methods for safe decision-making under uncertainties (e.g., Gaussian Process (GP)-MPC~\cite{ostafew2016learning}, GP-based robust control~\cite{berkenkamp2015safe}, neural adaptive control~\cite{joshi2019deep}, and learning-based CBF safety filter~\cite{taylor2020learning}). The core idea is enforcing convergence positive invariance of a set that can be the equilibrium in stabilization tasks~\cite{richards2018lyapunov}, or a state constraint set in more general settings~\cite{ostafew2016learning,taylor2020learning}. Whether using classical or more advanced control techniques for safe decision-making, theoretical guarantees are often provided for constraint sets that are designed a priori~\cite{brunke2022safe}. 

%There is a gap between what is perceived in practical settings and the theoretical results. 

% the safety constraints are often assumed to be known ahead of time validation in simulated setting and in well-structured environments.  
In practice, robots must perceive the environment based on high-dimensional sensory data and compute safe actions from perception. There are a few safe decision-making works that consider perception inputs. For instance, \cite{dean2021guaranteeing} proposes a measurement-robust CBF formulation to account for perception errors, \cite{xiao2023barriernet} leverages differentiable optimization tools to learn a CBF safety filter for collision avoidance in an end-to-end pipeline, \cite{grandia2023perceptive}~proposes a pipeline for safe locomotion based on plane segmentation, and \cite{Jian2022DynamicCB}~introduces a dynamic CBF for avoiding dynamic obstacles based on LiDAR measurements. The perception-based approaches, however, have not fully exploited object-level semantic and geometric information that can be obtained from a perception system. In this work, we aim to close the perception and action loop for safe navigation in semi-static environments.
% THere are also theoretical works  For instance, differentiable CBF, macro hutter paper, . 

% MPC, CBF for constraint satisfaction, explicit state and input constraints given

% this work, constraint directly built based on map (semantic label, stationarity score)

% combining MPC-CBF to leverage implicit prediction

% cluttered environments

\end{spacing}

%% file: sections/problem_formulation.tex
\vspace{-0.03\baselineskip}
\section{APPROACH OVERVIEW}
\label{sec:problem_formulation}
\begin{spacing}{0.97}
We explore the closed-loop perception and control problem in semi-static indoor environments, where the robot must simultaneously localize itself, maintain an up-to-date object-level map, track potential changes in the scene, and plan a safe trajectory to reach a desired target pose. This section presents an overview of our proposed approach. A high-level flow diagram of the approach is shown in Figure~\ref{fig:pipeline}. 

% \subsection{Problem Setup}

%In our setup, For an omnidirectional robot, the kinematics model $f$ has a linear form:

% \begin{equation}
% \label{eqn:robot-model-linear}
%     f(\mathbf{x}_t,\mathbf{u}_t) = \bA\mathbf{x}_t+ \bB\mathbf{u}_t.
% \end{equation}

%$f(\mathbf{x}_t,\mathbf{u}_t) = \bA\mathbf{x}_t+ \bB\mathbf{u}_t$. %the function $$. % $\bA$ is an identity matrix, and $\bB$ is a vector of ones.

% \subsection{Environment Assumptions}
\vspace{-0.1cm}
\subsection{Robot System}
We consider a ground robot that can be modelled as follows:
$$
    \mathbf{x}_{t+1} = 
    f(\mathbf{x}_t,\mathbf{u}_t),
$$
where $t\in\Z_{\ge 0}$ is the discrete-time index, $\bx=[x,y,\theta]^T\in \set{X}\subset \R^3$ is the 3-DoF global pose (position and orientation) of the robot with $\set{X}$ denoting the set of admissible states, $\mathbf{u}=[v_x,v_y,\omega]^T\in\ \mathcal{U}\subset \R^3$ is the desired linear and angular velocity in the world frame to the robot low-level controller with $\set{U}$ denoting the set of admissible inputs, $f:\set{X} \times \set{U} \mapsto \set{X}$ is a Lipschitz continuous function. 
%and $(\bA, \bB)$ are constant matrices of appropriate dimensions. 
In this work, we assume that the sets $\set{X}$ and $\set{U}$ are convex polytopic sets.
%In this work, we assume that the sets $\set{X}$ and $\set{U}$ are convex sets.

% =\bA\mathbf{x}_t+ \bB\mathbf{u}_t
% In this work, we assume that the robot model has a linear form:
% \begin{equation}
% \label{eqn:robot-model-linear}
%     f(\mathbf{x}_t,\mathbf{u}_t) = \bA\mathbf{x}_t+ \bB\mathbf{u}_t,
% \end{equation}
% which can be either derived from first principles or learned from data collected from the system.
\vspace{-0.05cm}
\subsection{Perception and Mapping}
The robot is expected to navigate around rigid objects in a semi-static scene. It is equipped with an RGB-D camera; at each timestep it outputs a frame, $\mathbf{F}_t$, that contains an RGB image and an aligned depth image. We further segment $\mathbf{F}_t$ into semantic pointcloud observations $\mathcal{Y}_{t} = \{\mathbf{Y}_{t,j}\}_{j=1 \dots J}$. Given $\mathcal{Y}_{t}$, the localization and mapping systems estimate the current robot pose, $\hat{\mathbf{x}}_t$, and update a mapped object library, $\mathcal{O}=\{\mathbf{O}_i\}_{i=1 \dots I}$, as discussed in Section~\ref{sec:scene_rep}-\ref{sec:bayesian_update}. 

We assume high-level knowledge of the objects is available. Each object belongs to a semantic class, $c \in \{1 \dots K\}$, with an associated stationarity class (likelihood of change), $s \in \{0,1\}$, where $s=1$ denotes a \textit{likely-static} object (e.g., wall), and $s=0$ denotes a \textit{likely-dynamic} object (e.g., chair).

\subsection{Safe Navigation}
% In this work, we consider a navigation problem, where the robot is expected to reach a prescribed target position. 
Given the tracked objects, $\mathcal{O}$, safety constraints are constructed based on the semantic and geometric information to ensure \textit{(i)} static obstacles are avoided and \textit{(ii)} conservative behaviours are achieved around inconsistent or \textit{likely-dynamic} objects. In our work, we encode the safety constraint as a CBF. Based on the CBF, the controller computes actions to minimize the distance to the target while adhering to the map-based safety constraints.
\vspace{-0.3cm}

 %CBF is a continuously differentiable scalar function defined over the $x-y$ plane of the environment $h \colon \mathbb{R}^2 \mapsto \mathbb{R}$. Based on the estimated robot state and observed objects, the MPC finds an optimal, safe path over the prediction horizon.

% illustrated in Figure xx, which is consisted of an omnidirectional ground vehicle. The kinematics is represented by the following model
% \begin{align}
%     xx
% \end{align}

\end{spacing}

%% file: sections/methodology.tex
\section{METHODOLOGY}
\label{sec:methodology}
\begin{spacing}{0.98}

\subsection{Scene Representation}
\label{sec:scene_rep}
As our system operates in evolving environments where objects can change over time, it is essential to construct the dense map at an object-level. Following POCD~\cite{QianChatrathPOCD}, a recent work on mapping in semi-static scenes, each object, $\mathbf{O}_i$, in the object library, $\mathcal{O}$, consists of the following elements:
\begin{itemize}
    \item a global 3D position, $\mathbf{p}_{i}$, and 1D heading, $\phi_{i}$,
    \item an inferred semantic class, $c_i$, and stationarity class, $s_i$
    \item a state distribution, $p(l_i,v_i)$, to model the object's current geometric change, $l_i\in\mathbb{R}$, and consistency (likelihood that the object has not changed), $v_i \in [0,1]$,
    \item a TSDF reconstruction, $\mathbf{M}_i$, in the global frame.
\end{itemize}

In our indoor test scenarios, objects are restricted to rotate around the $z$-axis. The object TSDF encodes the approximated distance in the global frame to the object surface, with positive values for points outside the object and negative values for points inside the object. All objects in $\mathcal{O}$ jointly represent the scene, and a trilinearly interpolated 3D TSDF map, $\mathcal{M}_{tsdf} \colon \mathbb{R}^3 \mapsto \mathbb{R}$, can be obtained by overlapping object-level TSDFs and taking the minimum at each voxel. 
\vspace{-0.1cm}
\subsection{Localization and Object-Level Mapping}
\label{sec:loc_map_overview}
We build our system on top of ORB-SLAM3~\cite{ORBSLAM3_TRO}, a state-of-the-art feature-based SLAM pipeline, and POCD~\cite{QianChatrathPOCD}. The system starts with an empty object library, $\mathcal{O} = \varnothing$. At each timestamp, the RGB-D frame, $\mathbf{F}_t$, and 3D observations, $\mathcal{Y}_t$, are input to our modified ORB-SLAM3 to estimate the current robot pose, $\hat{\mathbf{x}}_t$. To account for potential scene changes, only feature points belonging to \textit{likely-static} observations (e.g., walls) are used in the estimation.

With the estimated robot pose, $\hat{\mathbf{x}}_t$, the observations, $\mathcal{Y}_t$, are associated with the mapped objects in $\mathcal{O}$ using the Hungarian algorithm based on geometric and semantic consistency. Unmatched observations are used to spawn new objects. Each matched observation is integrated into the associated object, and the object's state model, $p(l,v)$, is propagated. Note that we employ a decoupled localization and mapping approach due to its high efficiency for real-time performance. A brief overview of the state propagation is provided in Section~\ref{sec:bayesian_update}, and more details of the mapping pipeline are available in~\cite{QianChatrathPOCD}.

\subsection{Object Consistency Estimation in Semi-Static Scenes} 
\label{sec:bayesian_update}

We follow POCD~\cite{QianChatrathPOCD} to propagate the object state model and estimate the consistency of potentially-changing objects, as it has shown to provide more consistent and noise-robust 3D mapping when faced with discrete scene changes. Specifically, each mapped object, $\mathbf{O}$, maintains a joint distribution, $p(l,v)$, where $v \in (0,1)$ models the likelihood of the object being consistent with historical observations, and $l\in\mathbb{R}$ models the magnitude of geometric change. In our implementation $l$ estimates the average TSDF change between the initial reconstruction and subsequent measurements. 

\iffalse
The state distribution is parametrized as a Gaussian-Beta product:
% \vspace{-0.1\baselineskip}
$$
p(l,v \mid \mu,\sigma,\alpha,\beta) \defeq \mathcal{N}(l \mid \mu, \sigma^2)\textrm{Beta}(v \mid \alpha, \beta)
$$
\noindent Intuitively, we expect the object to stay around its previously observed location, and the consistency is a probability between 0 and 1. 
\fi

When a new observation is available, a geometric measurement, $p\left(\Delta \mid l, v\right)$, and a semantic measurement, $p\left(s \mid v\right)$, are constructed. Here, $\Delta \in \mathbb{R}$ is the average TSDF difference between the object reconstruction and its current depth observation, and $s$ encodes our knowledge on the object stationarity (e.g., $s=1$ if the object is \textit{likely-static}, such as a table, and $s=0$ if the object is \textit{likely-dynamic}, such as a cart). These two measurements are used to update the object state model via a Bayesian update rule derived in~\cite{QianChatrathPOCD}.

\iffalse
Finally, an approximated, closed-form posterior, also in the form of a Gaussian-Beta product, can be obtained via moment matching. The exact update rule is available in~\cite{QianChatrathPOCD}:

$$
\begin{array}{l}
     q(l,v \mid \mu',\sigma',\alpha',\beta') \simeq\\
     \quad p(\Delta \mid l,v)p(s \mid v)p(l,v \mid \mu,\sigma,\alpha,\beta)
 \end{array}
$$
\fi
In this work, we use both the consistency, $\mathbb{E}[v]$, and stationarity label, $s$, to regulate the CBF around objects.

\subsection{Control Barrier Function (CBF)}
CBF certification~\cite{ames2019control} provides a means to guarantee the safety of a system in the sense of positive invariance of a state constraint set. Let $h$ be a continuously-differentiable function. Consider a constraint set, $\set{C}$, parameterized as the zero-superlevel set of $h$. The constraint set and its  boundary, $\partial\set{C}$, are defined as $\set{C} = \{\bx\in\set{X}\:|\: h(\bx)\ge 0\}$ and $\partial \set{C} = \{\bx\in\set{X}\:|\: h(\bx)= 0\}$, respectively.
%where $\set{X}\subset \R^n$ is the set of admissible states.
% \begin{align}
%     \set{C} &= \{x\in\R^n\:|\: h(x)\ge 0\}\\
%     \partial \set{C} &= \{x\in\R^n\:|\: h(x)\ge 0\}
% \end{align}

For a continuous-time system, a continuously-differentiable function, $h: \set{X}\mapsto \R$, is a CBF of the system if \textit{(i)} $\frac{\partial h}{\partial \bx} (\bx)\neq \mathbf{0}, \forall\bx\in\partial \set{C}$, and \textit{(ii)} there exists an extended class-$\set{K}_\infty$ function, $\gamma:\R\mapsto\R$, such that the following condition is satisfied~\cite{ames2019control}:
$$
    \sup_{\bu\in\R^m}\: \dot{h}(\bx,\bu) \ge -\gamma\left( h(\bx)\right),\hspace{1em} \forall \bx\in\set{X},
    \label{eqn:cbf_definition}
$$
where $\dot{h}(\bx,\bu) =\frac{\partial h}{\partial \bx}(\bx) \:\dot{\bx} =\frac{\partial h}{\partial \bx}(\bx)\: f_c(\bx,\bu)$ with $\dot{\bx} = f_c(\bx,\bu)$ being the continuous-time model of the system.

% where $\set{U}\subset\R^m$ denotes the set of admissible inputs~\cite{ames2019control}.

% The CBF framework has been commonly used for designing safety filters to certify learning-based controllers that do not have safety guarantees~\cite{brunke2022safe}. In these frameworks, it is usually assumed that a CBF $h(\bx)$ is given, and 
Given a CBF, the following condition is often imposed as a constraint in the safety filter design to guarantee positive invariance of the constraint set:
$$
    \dot{h}(\bx,\bu) \ge -\gamma\left( h(\bx)\right).
$$
\vspace{-1.5\baselineskip}
 
 %This idea of CBF is closely related to Nagumo's theory~\cite{nagumo1942lage}. In particular, it enforces a non-negative Lie derivative of $h$ at the boundary $\partial\set{C}$ to guarantee positive invariance. When the system is inside of the constraint set, the Lie derivative of $h$ is lower bounded, usually by a negative number; this lower bound determines how fast the system is allowed to approach the boundary $\partial\set{C}$ from the interior. 

 For a discrete-time system, a similar CBF condition can be defined as follows:
$$
     \Delta h(\bx_t,\bu_t) \ge -\bar{\gamma} h(\bx_t),
$$
where $ \Delta h(\bx_t,\bu_t)  =   h\big(f(\bx_t, \bu_t)\big)- h(\bx_{t})$, and  $\bar{\gamma} \in (0,1]$ is a constant scalar~\cite{zeng2021safety}. 

\subsection{Bridging Object Mapping and CBFs}
\label{sec:construct_cbf}

\subsubsection{Non-Semantic CBF Construction}
\label{sec:non_sem_cbf}
 The object TSDFs used to construct the global map, $\mathcal{M}_{tsdf}$, are based on ray-tracing from estimated robot poses, causing bias around object boundaries with limited views. Following prior work~\cite{Jian2022DynamicCB,Singletary2020ComparativeAO,Han2019FIESTAFI}, we generate a 2.5D non-semantic Euclidean Distance Function (EDF) map, denoted as $\Tilde{\mathcal{M}}_{edf} \colon \mathbb{R}^2 \mapsto \mathbb{R}$. To capture the obstacles within the robot's traversal space, we first extract the TSDF layers between the ground plane and a specified height threshold, $\theta_{\textrm{z}}$. We concatenate the layers into a single 2.5D map by taking the minimum of the absolute value along the $z-$axis. The approximated zero-level set, $\partial\Tilde{\mathcal{M}}_{edf}$, thresholded by a small value, $\theta_{\textrm{zero}}$, representing the obstacle surfaces, is extracted, and the Euclidean distance is computed for $ \Tilde{\mathcal{M}}_{edf}$. The process is formalized as follows:
% \vspace{-0.6\baselineskip}
\begin{subequations}
\label{eq:raw_esdf}
\begin{align}
    & \mathcal{M}_{2.5D}(x,y) = \min_{0 < z \leq  \theta_{z}} |\mathcal{M}_{tsdf}(x,y,z)| \label{eq:raw_esdf_1}\\
    & \partial\Tilde{\mathcal{M}}_{edf} = \{(x,y) \mid \mathcal{M}_{2.5D}(x,y) \leq \theta_{\textrm{zero}} \} \label{eq:raw_esdf_2}\\
    & \Tilde{\mathcal{M}}_{edf}(x,y) = \min_{(\bar{x},\bar{y})\in \partial\Tilde{\mathcal{M}}_{edf}} \|(x,y)  - (\bar{x},\bar{y})\|_2 \label{eq:raw_esdf_3}
\end{align}
\end{subequations}
\vspace{-0.5\baselineskip}

We apply two transformations to encode $\Tilde{\mathcal{M}}_{edf}$ into a CBF. To account for the size of the robot, we apply a negative bias, $b$, which introduces a buffer around the obstacles. To ensure the robot's behaviour remains unaffected when far from obstacles, we truncate $\Tilde{\mathcal{M}}_{edf}$ by a cutoff threshold, $\theta_{\textrm{cutoff}}$. Formally, the continuous non-semantic CBF, denoted by $\Tilde{h}$, is defined as:
% \vspace{-0.1\baselineskip}
\begin{equation}
\label{eq:non_sem_cbf}
    \Tilde{h}(\mathbf{x}) = \min \{\Tilde{\mathcal{M}}_{edf}(x,y)-b, \theta_{\textrm{cutoff}} \}
\end{equation}
\noindent In our implementation, the continuity is achieved via bilinear interpolation over a discrete 2D grid, and gradients of $\Tilde{h}$ is obtained via finite difference.

\begin{figure}[t]
\centering
  \includegraphics[width=\linewidth]{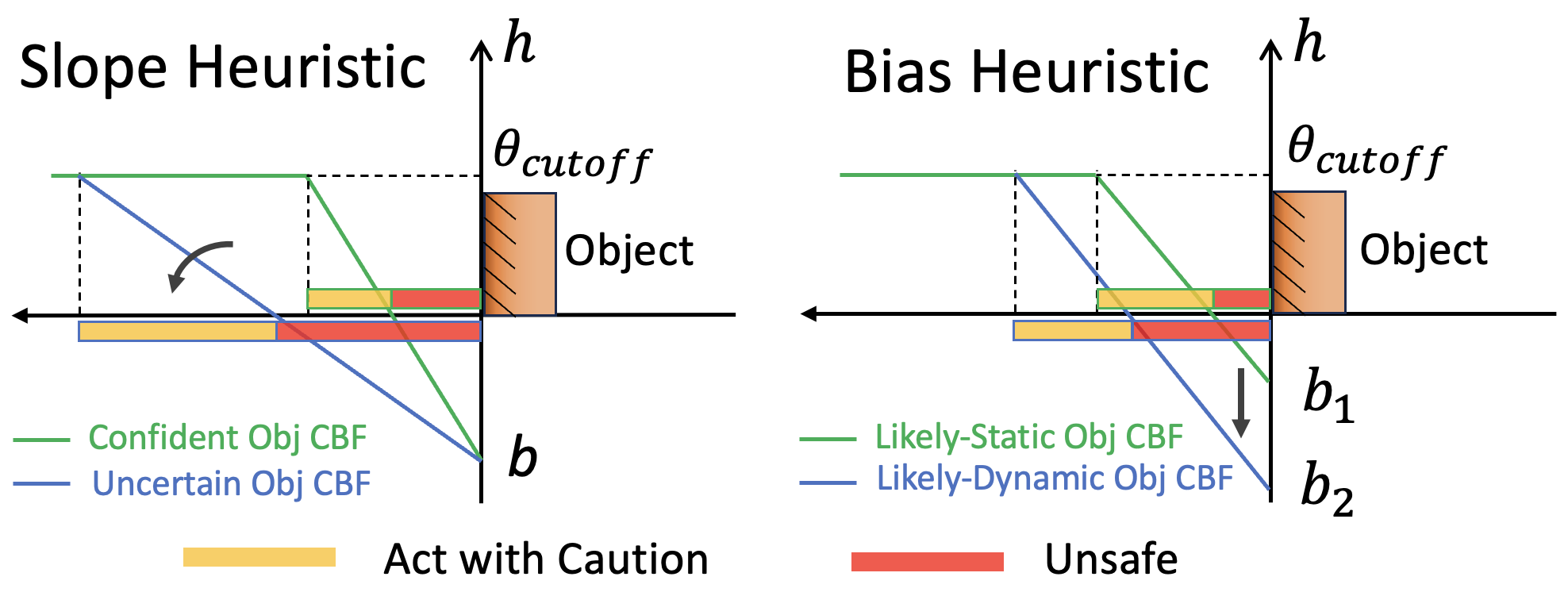}
\caption{Visualization of the proposed heuristics for CBFs around objects. The Slope Heuristic scales down the slope of the CBF for uncertain objects, inflating both the unsafe region (red) and the cautious region (orange), causing the robot to act more conservatively. The Bias Heuristic only inflates the unsafe region around \textit{likely-dynamic} objects to create a larger safety buffer, and the robot's behavior is unaffected outside. The two heuristics can be combined to achieve the desired behaviours.}
\vspace{-1.2\baselineskip}
\label{fig:heuristics}
\end{figure}

\subsubsection{Incorporating Semantic and Geometric Knowledge} The previously defined CBF, $\Tilde{h}$, does not consider any object-level semantics and consistency knowledge, which may lead to risky behaviour such as moving too close to potentially changed objects. We propose the Slope Heuristic and the Bias Heuristic to generate an object-aware EDF map, $\mathcal{M}_{edf}$, for enhanced safety and adaptability. We first augment the zero-level set, $\partial\Tilde{\mathcal{M}}_{edf}$, with the the object consistency, $\mathbb{E}[v]$, and stationarity class, $s$, for which the voxel belongs to:

\iffalse
\begin{equation}
\label{eq:heur_1}
\begin{aligned}
    & \partial\mathcal{M}_{edf} \\
    & = \{(x,y, \mathbb{E}[v_i], s_i) \mid \mathcal{M}_{2.5D}(x,y) \leq \theta_{\textrm{zero}}, (x,y) \in \mathbf{O}_i \} \\
\end{aligned}
\end{equation}
\vspace{-1\baselineskip}
\fi

\vspace{-0.9\baselineskip}
\begin{equation}
\begin{aligned}
     &\partial\mathcal{M}_{edf} \\
    = &\{(x,y, \mathbb{E}[v_i], s_i) \mid \mathcal{M}_{2.5D}(x,y) \leq \theta_{\textrm{zero}}, (x,y) \in \mathbf{O}_i \} 
\end{aligned}
\end{equation}
\vspace{-0.3\baselineskip}

The Slope Heuristic scales the Euclidean distances in (\ref{eq:raw_esdf_3}) to regulate both the permissible proximity of the robot to uncertain objects and the aggressiveness when navigating around them. Under the common bias, $b$, and cutoff threshold, $\theta_{\textrm{cutoff}}$, CBFs with lower slopes will have inflated unsafe regions, and the robot will behave more conservatively when approaching them. We propose to utilize the adjusted object-level consistency estimate, $\lambda_{c}\mathbb{E}[v]$, as the scaling factor, where $\lambda_{c}$ is a tunable parameter. 
\newpage
The Bias Heuristic changes the bias to solely control the unsafe region around objects. We propose to upscale $b$ with a tunable parameter, $\lambda_{s} > 1$, for objects that are \textit{likely-dynamic} (s=0). This ensures that the robot does not get too close to potentially changing objects, but can also move freely, when having a high level of confidence. Combining the two heuristics, our object-aware CBF, $h$, is formulated as
\begin{equation}
\begin{aligned}
    &\mathcal{M}_{edf}(x,y)\\
    =&\min_{(\bar{x},\bar{y},\mathbb{E}[v],s)\in \partial\mathcal{M}_{edf}} \lambda_{c}\mathbb{E}[v]\|(x,y)  - (\bar{x},\bar{y})\|_2 \\
    &\hspace{3.5cm} - [\lambda_s(1-s) + s] b
\end{aligned}
\end{equation}
\vspace{-0.15\baselineskip}
\begin{equation}
h(\mathbf{x}) = \min \{\mathcal{M}_{edf}(x,y), \theta_{\textrm{cutoff}} \}
\end{equation}
We visualize the effect of the two heuristics in Figure~\ref{fig:heuristics}.

\iffalse
\begin{figure}
    \centering
    \includegraphics[width=\columnwidth]{./figures/mpc-and-mpc-cbf.png}
    \caption{An illustration of the difference between MPC-CBF and a typical MPC. The CBF implicitly captures the desired behaviour and allows the robot to steer away from the unsafe region ahead of time. For a typical MPC with state constraints, this behaviour is only achievable with a sufficiently long prediction horizon, which can be computationally expensive~\cite{zeng2021safety}. }
    \label{fig:mpc-and-mpc-cbf}
\vspace{-1.0\baselineskip}
\end{figure}
\fi
\vspace{-0.25\baselineskip}
\subsection{Discrete-Time Safe Control}
\label{sec:mpc_cbf}

In this work, inspired by~\cite{zeng2021safety}, we use an MPC-CBF framework for computing safe actions based on the object-level map. At each time, $t$, given the current state, $\bx_t$, and the CBF, $h$, the following optimization problem is solved:
\begin{subequations}
\label{eq:mpc_cbf_opt}
\begin{align}
&\hspace{-0.6em}\min_{\bx_{t+1:t+T|\:t}\atop\: \bu_{t:t+T-1|\:t}}\hspace{0.2em} \sum_{k=0}^{T-1} l_t\left ( \bx_{t+k\:|\:t},\bu_{t+k\:|\:t} \right ) + l_T\left (\bx_{t+T\:|\:t}\right )\label{subeq:mpc-cost}\\
&\hspace{-0.3em}\text{s.t.}\hspace{0.6em} \bx_{t\:|\:t} = \bx_{t},\label{subeq:mpc-initial}\\
&\hspace{1.5em}  \bx_{t+k+1\:|\:t} = f(\bx_{t+k\:|\:t}, \bu_{t+k\:|\:t}), \, \forall k\in\set{T}, \label{subeq:robot-model-in-mpc}\\
&\hspace{1.5em}   \bx_{t+k\:|\:t} \in \set{X}, \:\bu_{t+k\:|\:t} \in \set{U}, \, \forall k\in\set{T},\label{subeq:mpc-asets}\\
&\hspace{1.5em} \Delta h(\bx_{t+k\:|\:t},\bu_{t+k\:|\:t}) \ge -\bar{\gamma}\: h(\bx_{t+k\:|\:t}), \, \forall k\in\set{T},
\label{subeq:cbf-constraint-in-mpc}
\end{align}
\label{eqn:mpc-cbf}%
\end{subequations}
where $T$ is the prediction horizon length, $\set{T} = \{0,1,..., T-1\}$ is the set of time indices over the prediction horizon, ``:'' in the subscripts denotes consecutive timesteps, and $l_t$ and $l_T$ are the stage cost and the terminal cost, respectively. 
%$\Delta h(x_{t+k+1\:|\:t}) = h(x_{t+k\:|\:t})$, and $f$ is the robot dynamics or kinematics model. 
After solving the optimization problem in~\eqref{eqn:mpc-cbf}, the first input, $\bu_{t\:|\:t}$, is applied to the system.

% The MPC-CBF framework allows us to predict forward based on the current map and directly encode safety constraints in the CBF condition. 
In contrast to a typical MPC, the MPC-CBF framework allows us to \textit{(i)} conveniently define the desired behaviour of the robot via a CBF directly derived based on map information, and \textit{(ii)} reduces the prediction horizon required to achieve a desired level of safety behaviour~\cite{zeng2021safety}.
%The MPC-CBF framework allows us to predict forward based on the current map and directly encode safety constraints in the CBF condition. In contrast to a typical MPC framework, the incorporation of the CBF allows us to \textit{(i)} conveniently define the desired behaviour of the robot directly based on map information, and \textit{(ii)} reduces the prediction horizon required to achieve a desired level of safety behaviour~\cite{zeng2021safety}. %(see illustration \autoref{fig:mpc-and-mpc-cbf}).

Note that the MPC-CBF is not a standard convex optimization problem due to the dynamics in~\eqref{subeq:robot-model-in-mpc} and the CBF constraint in~\eqref{subeq:cbf-constraint-in-mpc}. To allow fast online computation, we approximate~\eqref{subeq:robot-model-in-mpc} and~\eqref{subeq:cbf-constraint-in-mpc} by linearizing them about the prediction trajectory from the previous timestep. 
% To allow fast online computation, we use a sequential convex programming (SQP) approach where the constraint  in~\eqref{subeq:cbf-constraint-in-mpc} is replaced by linearizing the condition about the prediction trajectory from the previous timestep. 
For each timestep, $k\in\set{T}$, we introduce a new set of decision variables, $\delta\bx_{t+k\:|\:t} = \bx_{t+k\:|\:t} - \bx_\text{op}$ and $\delta\bu_{t+k\:|\:t} = \bu_{t+k\:|\:t} - \bu_\text{op}$, where $\bx_\text{op}=\bx_{t-1+k\:|\:t-1}$ and $\bu_\text{op} = \bu_{t-1+k\:|\:t-1}$.
The linearized dynamics and CBF constraints can be written in  forms that are affine in the decision variables:
\vspace{-0.1cm}
\begin{align}
 & \delta\bx_{t+k+1|t} = \bA(\bx_\text{op}, \bu_\text{op})\hspace{0.01em}\delta\bx_{t+k|t} + \bB \hspace{0.01em}(\bx_\text{op}, \bu_\text{op})\delta\bu_{t+k|t},\label{eqn:linearized-model}\\
&   \bC(\bx_\text{op},\bu_\text{op})\hspace{0.01em} \delta\bx_{t+k|t}  + \bD(\bx_\text{op},\bu_\text{op})\hspace{0.01em}\delta\bu_{t+k|t} +  c(\bx_\text{op},\bu_\text{op})\ge 0,\,
    \label{eqn:linearization}
\end{align}
where~$c$, $\bA$, $\bB$, $\bC$, and $\bD$ are constant terms derived from the linear approximations. In our implementation, \eqref{eqn:linearization} requires querying $h$ and its gradient, $\frac{\partial h}{\partial \bx}$, at $\{\bx^*_{t\:|\:t-1},..., \bx^*_{t-1+T\:|\:t-1}\}$.
%which is affine in the decision variables. 
By replacing the dynamics,~\eqref{subeq:robot-model-in-mpc}, with~\eqref{eqn:linearized-model}, and the CBF constraint,~\eqref{subeq:cbf-constraint-in-mpc}, with~\eqref{eqn:linearization}, and re-expressing \eqref{subeq:mpc-cost}, \eqref{subeq:mpc-initial}, and \eqref{subeq:mpc-asets} in terms of $\delta \bx$ and $\delta \bu$, we obtain a quadratic program that can be solved efficiently. %The input applied to the system is $\bu_{t-1\:|\:t-1} + \delta \bu_{t\:|\:t}$.  %In our case, we consider a model of the robot that has a linear form~\eqref{eqn:robot-model-linear}; if $f(\bx_t, \bu_t)$ is nonlinear, a similar approximation scheme can be used.

% The problem in~\eqref{eqn:mpc-cbf} is not a standard quadratic program (QP) as in other works. 
% Even though the robot kinematics model is linear and control affine, the CBF constraint is nonlinear. We approximate with linearization about the previous prediction sequence. This requires the h values and gradient information at xx 

% write about how tracking error is accounted for, similar to the food preparation paper

\end{spacing}

%% file: sections/exp_results.tex
\begin{spacing}{0.97}
\section{Simulation Results}

\subsection{Experiment Setup}
\label{sec:sim_setup}

We demonstrate the key features of our object-aware MPC-CBF framework in simulation. As seen in Figure~\ref{fig:cover}, an indoor scenario consisting of multiple \textit{walls} and one \textit{drawer} is set up in PyBullet~\cite{pybullet}. The \textit{walls} are considered \textit{likely-static}. An omnidirectional robot, modeled as single integrators in each dimension, is simulated and controlled by our system. The robot is 50 cm in diameter and equipped with a RGB-D sensor that outputs color images, Gaussian corrupted depth measurements, and ground-truth semantic masks. To construct the object-aware CBF, we set the zero-level threshold, $\theta_{\textrm{zero}}$, to 0.15 m, truncation distance, $\theta_{\textrm{cutoff}}$, to 1.8 m, bias, $b$, to 0.75 m, consistency factor, $\lambda_c$, to 3.0, and stationarity factor, $\lambda_s$, to 2.0.

\subsection{Object Semantic Knowledge and Consistency}

\begin{figure}[t]
\centering
  \includegraphics[width=1\linewidth]{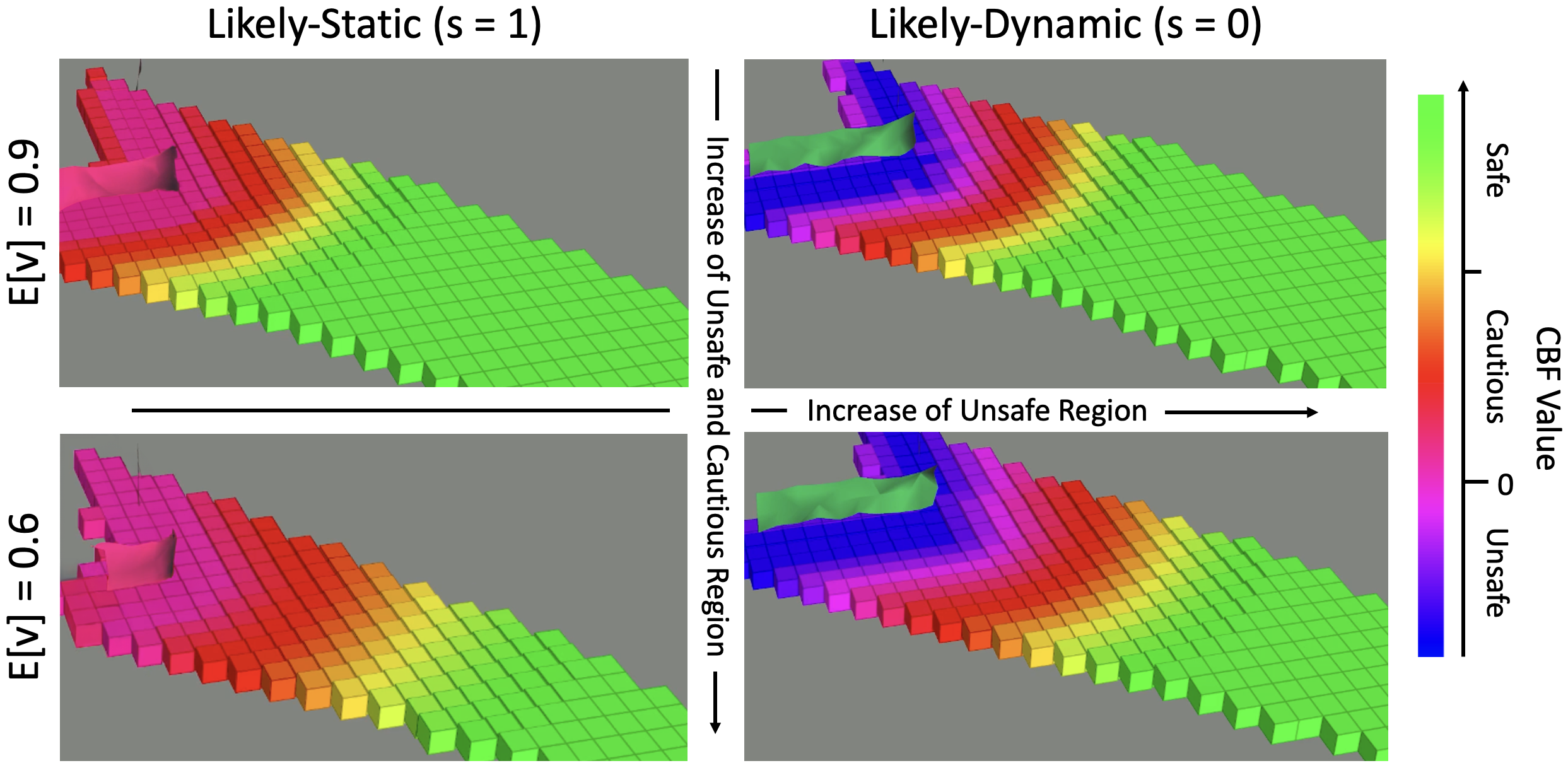}
\caption{Visualization of the object-aware CBFs around the \textit{drawer} object under different stationarity labels and consistency estimates. }
\vspace{-1.0\baselineskip}
\label{fig:sim_exp_consist}
\end{figure}

We validate our CBF formulation's capture of object-level semantic knowledge and consistency estimates, as shown in Figure~\ref{fig:sim_exp_consist}. As the robot approaches the stationary \textit{drawer}, its consistency estimate increases with more observations, causing the unsafe (enclosed by pink grids) and the cautious (enclosed by yellow grids) regions to contract. When testing the \textit{drawer} with a \textit{likely-dynamic} label, the Bias Heuristic effectively increases the unsafe region, while the cautious region remains the same (under the same consistency level) due to the unchanged slope of the CBF.

\subsection{Balancing Feasibility and Conservatism}

\begin{figure}[t]
\centering
  \includegraphics[width=0.8 \linewidth]{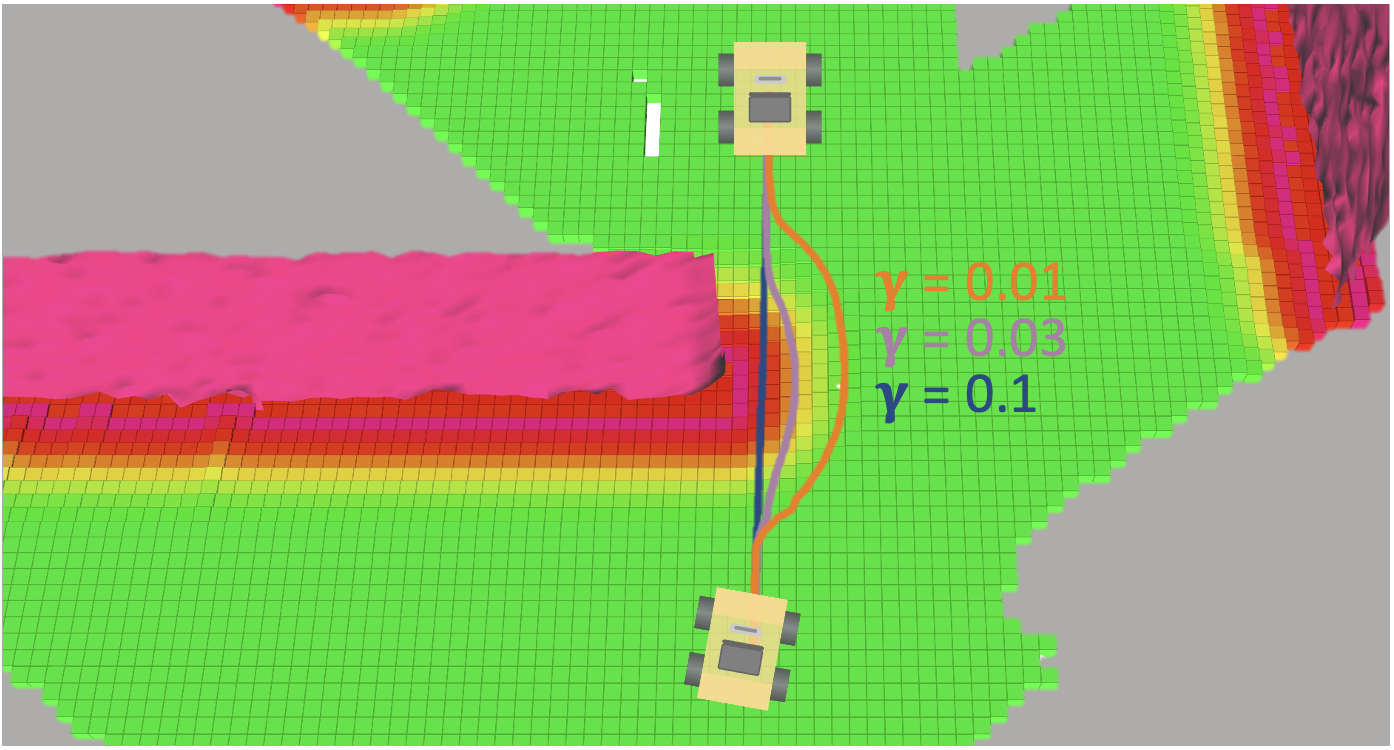}
\caption{Ground-truth robot trajectories with different values of $\bar{\gamma}$ in the MPC-CBF safety constraint (\ref{subeq:cbf-constraint-in-mpc}).}
\vspace{-1\baselineskip}
\label{fig:sim_exp_gamma}
\end{figure}

The scalar, $\bar{\gamma}$, in the MPC-CBF safety constraint, (\ref{subeq:cbf-constraint-in-mpc}), is a critical design variable. A smaller $\bar{\gamma}$ makes the controller safer but may render the optimization in (\ref{eq:mpc_cbf_opt}) infeasible. Conversely, a larger $\bar{\gamma}$ improves controller feasibility but elevates risk. Figure~\ref{fig:sim_exp_gamma} shows robot trajectories under different $\bar{\gamma}$ choices, with lower values resulting in more conservative behaviour around the wall. Note that with $\bar{\gamma}=0.1$, the robot went straight through the red cautious region but did not hit the pink zero-level set. In real-world experiments, we opt for $\bar{\gamma}=0.03$ so the system is not overly conservative under measurement noise and localization errors. 

\section{Real World Experiments}

\subsection{Experiment Setup}
We demonstrate the closed-loop capability of our system on a real-world robot. As in the simulation tests, we use an omnidirectional mobile base equipped with an Intel Realsense D435 RGB-D camera for perception. The pipeline runs on an external laptop with an Intel i7-8850H CPU, at a fixed rate of 5 Hz. A low-level onboard PID controller follows the MPC trajectory at 20 Hz.  We set up the test scenario as seen in Figure~\ref{fig:real_exp_sem} and adopted the same parameter settings from Section~\ref{sec:sim_setup}. While we tested on an omnidirectional robot, our method can be generalized to other kinematic models.
\vspace{-0.1cm}
\subsection{Obstacle Avoidance with Semantic Knowledge}
In this experiment, the robot navigates through a narrow gap between a wheeled drawer (\textit{likely-dynamic}, left), and a flat-base drawer (\textit{likely-static}, right). We compare our object-aware MPC-CBF against two baselines derived from (\ref{eq:mpc_cbf_opt}): a naive MPC-CBF using the non-semantic CBF $\Tilde{h}$ in (\ref{eq:non_sem_cbf}), and a classic MPC replacing the full CBF constraint (\ref{subeq:cbf-constraint-in-mpc}) with geometric state constraints, $\Tilde{h}(\mathbf{x}_{t+k\mid t})~>~0$. 

\begin{figure}[b]
\centering
  \vspace{-0.3cm}
  \includegraphics[width=\linewidth]{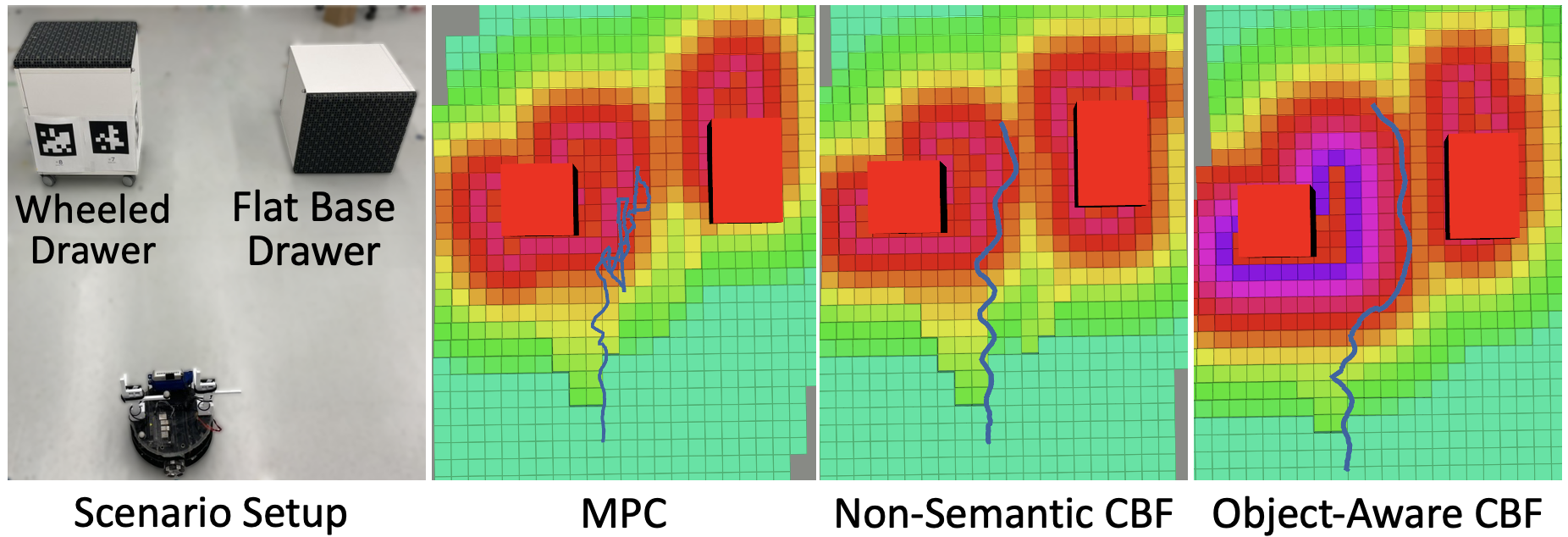}
\caption{The scenario (left) and visualization (right) of the ground-truth robot trajectories using three controllers. The wheeled drawer is considered to be likely to change ($s=1)$ while the flat base drawer is likely to stay ($s=0)$. The robot distinguishes between the two drawers based on color.}
\label{fig:real_exp_sem}
\end{figure}

Figure~\ref{fig:real_exp_sem} depicts the ground-truth trajectories recorded by a Vicon motion capture system. The classic MPC solely considers the state constraint at the unsafe region's boundary. Due to depth measurement noise and localization errors, the unsafe region frequently expands and contracts. As a result, the robot under the MPC often enters infeasible regions, getting stuck between the two drawers. 

In contrast, the non-semantic MPC-CBF successfully navigates the gap. Since the controller does not encode any semantic information, the robot opts for the middle path, disregarding potential risks from scene changes.

Our object-aware MPC-CBF recognizes that the left drawer may undergo changes, resulting in larger unsafe and cautious regions around it. Consequently, the robot navigates the gap while remaining closer to the \textit{likely-static} drawer. Although the scenario is simple, our local planning method can be scaled to more complex settings when combined with a global planner such as~\cite{Karaman2011AnytimeMP}.
\vspace{-0.1cm}
\subsection{Reacting to Scene Changes}

\begin{figure}[t]
\centering
  \includegraphics[width=0.9\linewidth]{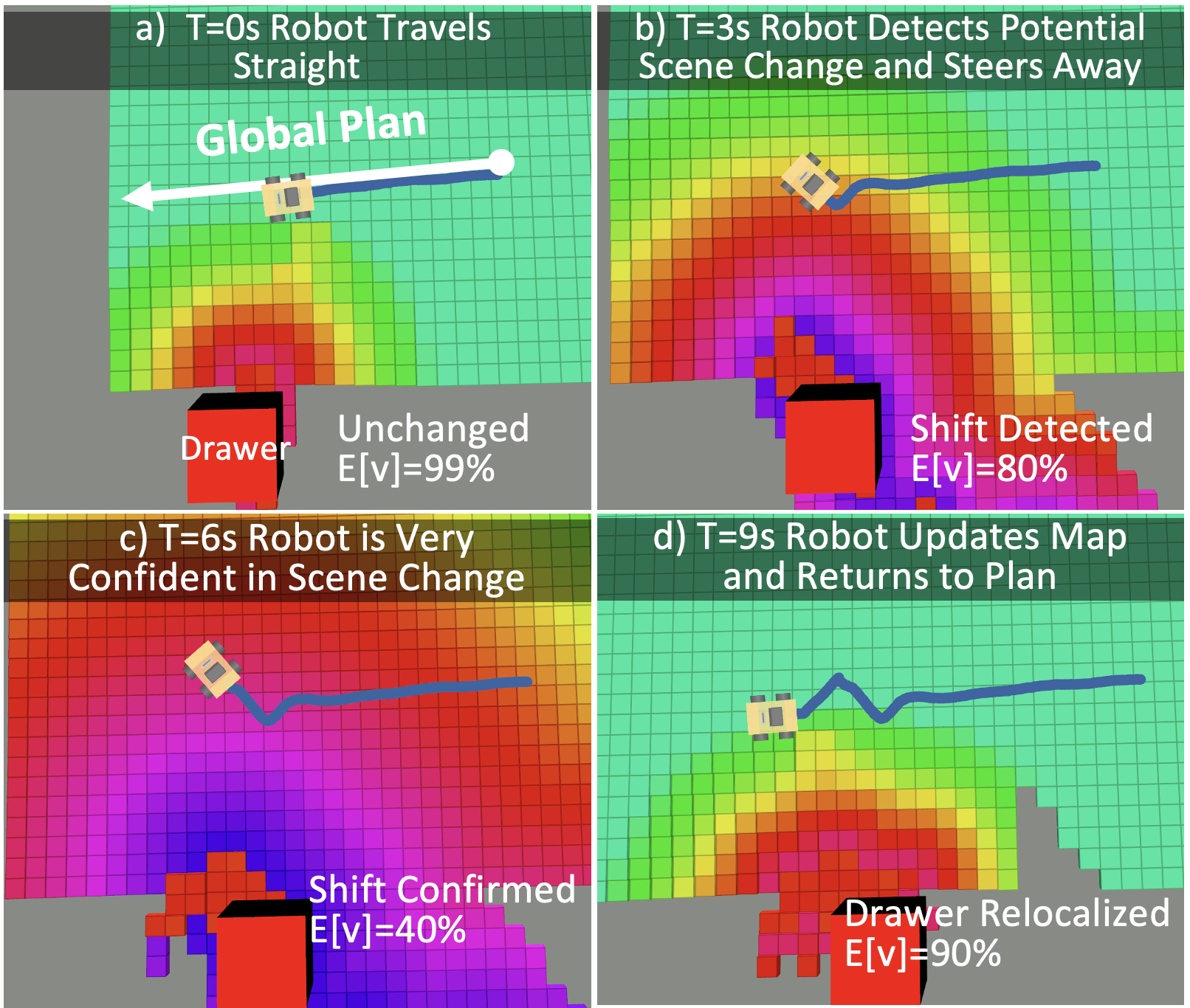}
\caption{An illustration of how our closed-loop system deviates from the original path in response to a sudden change in the scene.}
\vspace{-1\baselineskip}
\label{fig:real_exp_change}
\end{figure}

In the final experiment, we showcase the system's adaptability to sudden scene changes, as depicted in Figure~\ref{fig:real_exp_change}. Initially, the robot follows a straight path near a drawer (a). When the drawer is pushed, causing a scene change, our semi-static mapping system quickly detects it. The drawer's consistency estimate begins to drop, leading to an expansion of the unsafe region around it (b). The object-aware MPC-CBF responds to this by redirecting the robot away from its initial path (c). Once the drawer's consistency estimate falls below 40\%, it is removed from the object library and reconstructed at its new pose with high consistency. As a result, the unsafe region returns to normal, and the robot resumes its initial path, continuing towards its goal.

\end{spacing}

%% file: sections/conclusion.tex
\begin{spacing}{0.98}
In this work, we present a system designed for closed-loop perception and safe control within semi-static environments. Our approach involves encoding object-level semantic prior knowledge and consistency estimates into the CBF safety constraints within the MPC-CBF framework. This integration of scene semantics and consistency offers a versatile means to directly shape the robot's behavior in complex environments based on map-derived information. Through a series of experiments conducted in both simulated and real-world scenarios, we demonstrate how our system dynamically adapts its actions to address potential risks in the environment, thereby ensuring safe navigation. By addressing the challenges posed by changing environments, our work presents promising prospects for the development of robust systems capable of navigating extended periods in dynamic real-world settings.
\end{spacing}

%% file: root.bbl
\begin{thebibliography}{10}
\providecommand{\url}[1]{#1}
\csname url@samestyle\endcsname
\providecommand{\newblock}{\relax}
\providecommand{\bibinfo}[2]{#2}
\providecommand{\BIBentrySTDinterwordspacing}{\spaceskip=0pt\relax}
\providecommand{\BIBentryALTinterwordstretchfactor}{4}
\providecommand{\BIBentryALTinterwordspacing}{\spaceskip=\fontdimen2\font plus
\BIBentryALTinterwordstretchfactor\fontdimen3\font minus
  \fontdimen4\font\relax}
\providecommand{\BIBforeignlanguage}[2]{{%
\expandafter\ifx\csname l@#1\endcsname\relax
\typeout{** WARNING: IEEEtran.bst: No hyphenation pattern has been}%
\typeout{** loaded for the language `#1'. Using the pattern for}%
\typeout{** the default language instead.}%
\else
\language=\csname l@#1\endcsname
\fi
#2}}
\providecommand{\BIBdecl}{\relax}
\BIBdecl

\bibitem{Cadena2016PastPA}
\BIBentryALTinterwordspacing
C.~Cadena, L.~Carlone, H.~Carrillo, Y.~Latif, D.~Scaramuzza, J.~Neira, I.~D.
  Reid, and J.~J. Leonard, ``Past, present, and future of simultaneous
  localization and mapping: Toward the robust-perception age,'' \emph{IEEE
  Transactions on Robotics}, vol.~32, pp. 1309--1332, 2016. [Online].
  Available: \url{https://api.semanticscholar.org/CorpusID:9686224}
\BIBentrySTDinterwordspacing

\bibitem{panoptictsdf}
L.~M. Schmid, J.~A. Delmerico, J.~L. Sch{\"{o}}nberger, J.~I. Nieto,
  M.~Pollefeys, R.~Siegwart, and C.~Cadena, ``Panoptic multi-{TSDF}s: a
  flexible representation for online multi-resolution volumetric mapping and
  long-term dynamic scene consistency,'' \emph{ICRA}, 2022.

\bibitem{QianChatrathPOCD}
J.~Qian, V.~Chatrath, J.~Yang, J.~Servos, A.~Schoellig, and S.~L. Waslander,
  ``{POCD: Probabilistic Object-Level Change Detection and Volumetric Mapping
  in Semi-Static Scenes},'' in \emph{2022 Robotics: Science and Systems (RSS)},
  2022.

\bibitem{Fu2023NeuSENS}
J.~Fu, Y.~Du, K.~Singh, J.~B. Tenenbaum, and J.~J. Leonard, ``Neuse: Neural
  se(3)-equivariant embedding for consistent spatial understanding with
  objects,'' in \emph{Robotics: Science and Systems (RSS)}, 2023.

\bibitem{QianPOVSLAM}
J.~Qian, V.~Chatrath, J.~Servos, A.~Mavrinac, W.~Burgard, S.~L. Waslander, and
  A.~Schoellig, ``{POV-SLAM: Probabilistic Object-Level Variational SLAM},'' in
  \emph{2023 Robotics: Science and Systems (RSS)}, 2023.

\bibitem{brunke2022safe}
L.~Brunke, M.~Greeff, A.~W. Hall, Z.~Yuan, S.~Zhou, J.~Panerati, and A.~P.
  Schoellig, ``Safe learning in robotics: From learning-based control to safe
  reinforcement learning,'' \emph{Annual Review of Control, Robotics, and
  Autonomous Systems}, vol.~5, pp. 411--444, 2022.

\bibitem{rosolia2017learning}
U.~Rosolia and F.~Borrelli, ``Learning model predictive control for iterative
  tasks. a data-driven control framework,'' \emph{IEEE Transactions on
  Automatic Control}, vol.~63, no.~7, pp. 1883--1896, 2017.

\bibitem{ostafew2016learning}
C.~J. Ostafew, A.~P. Schoellig, T.~D. Barfoot, and J.~Collier, ``Learning-based
  nonlinear model predictive control to improve vision-based mobile robot path
  tracking,'' \emph{Journal of Field Robotics}, vol.~33, no.~1, pp. 133--152,
  2016.

\bibitem{gahlawat2020l1}
A.~Gahlawat, P.~Zhao, A.~Patterson, N.~Hovakimyan, and E.~Theodorou, ``{L1-GP}:
  {L1} adaptive control with bayesian learning,'' in \emph{Proc. of the
  Learning for Dynamics and Control Conference}, 2020, pp. 826--837.

\bibitem{chowdhary2014bayesian}
G.~Chowdhary, H.~A. Kingravi, J.~P. How, and P.~A. Vela, ``Bayesian
  nonparametric adaptive control using {G}aussian processes,'' \emph{IEEE
  Transactions on Neural Networks and Learning Systems}, vol.~26, no.~3, pp.
  537--550, 2014.

\bibitem{ames2019control}
A.~D. Ames, S.~Coogan, M.~Egerstedt, G.~Notomista, K.~Sreenath, and P.~Tabuada,
  ``Control barrier functions: Theory and applications,'' in \emph{Proc. of the
  European control conference (ECC)}, 2019, pp. 3420--3431.

\bibitem{wang2018safe}
L.~Wang, E.~A. Theodorou, and M.~Egerstedt, ``Safe learning of quadrotor
  dynamics using barrier certificates,'' in \emph{Proc. of the IEEE
  International Conference on Robotics and Automation (ICRA)}, 2018, pp.
  2460--2465.

\bibitem{xiao2023barriernet}
W.~Xiao, T.-H. Wang, R.~Hasani, M.~Chahine, A.~Amini, X.~Li, and D.~Rus,
  ``Barriernet: Differentiable control barrier functions for learning of safe
  robot control,'' \emph{IEEE Transactions on Robotics}, 2023.

\bibitem{grandia2023perceptive}
R.~Grandia, F.~Jenelten, S.~Yang, F.~Farshidian, and M.~Hutter, ``Perceptive
  locomotion through nonlinear model-predictive control,'' \emph{IEEE
  Transactions on Robotics}, 2023.

\bibitem{ORBSLAM3_TRO}
C.~Campos, R.~Elvira, J.~J.~G. Rodríguez, J.~M. M.~Montiel, and J.~D.~Tardós,
  ``{ORB-SLAM3}: An accurate open-source library for visual, visual-inertial
  and multi-map {SLAM},'' \emph{IEEE Transactions on Robotics}, vol.~37, no.~6,
  pp. 1874--1890, 2021.

\bibitem{Pfister2000SurfelsSE}
\BIBentryALTinterwordspacing
H.~Pfister, M.~Zwicker, J.~van Baar, and M.~H. Gross, ``Surfels: surface
  elements as rendering primitives,'' \emph{Proceedings of the 27th annual
  conference on Computer graphics and interactive techniques}, 2000. [Online].
  Available: \url{https://api.semanticscholar.org/CorpusID:1756681}
\BIBentrySTDinterwordspacing

\bibitem{Bloesch2019LearningMF}
\BIBentryALTinterwordspacing
M.~Bloesch, T.~Laidlow, R.~Clark, S.~Leutenegger, and A.~J. Davison, ``Learning
  meshes for dense visual slam,'' \emph{2019 IEEE/CVF International Conference
  on Computer Vision (ICCV)}, pp. 5854--5863, 2019. [Online]. Available:
  \url{https://api.semanticscholar.org/CorpusID:203600671}
\BIBentrySTDinterwordspacing

\bibitem{dense_vslam_rgbd}
C.~Kerl, J.~Sturm, and D.~Cremers, ``Dense visual slam for rgb-d cameras,'' in
  \emph{2013 IEEE/RSJ International Conference on Intelligent Robots and
  Systems}, 2013, pp. 2100--2106.

\bibitem{octomap}
A.~Hornung, K.~M. Wurm, M.~Bennewitz, C.~Stachniss, and W.~Burgard,
  ``{OctoMap}: An efficient probabilistic {3D} mapping framework based on
  octrees,'' \emph{Autonomous Robots}, 2013.

\bibitem{voxblox}
H.~{Oleynikova}, Z.~{Taylor}, M.~{Fehr}, R.~{Siegwart}, and J.~{Nieto},
  ``Voxblox: Incremental 3d euclidean signed distance fields for on-board mav
  planning,'' in \emph{2017 IEEE/RSJ International Conference on Intelligent
  Robots and Systems (IROS)}, 2017.

\bibitem{slam++}
R.~F. Salas-Moreno, R.~A. Newcombe, H.~M. Strasdat, P.~H.~J. Kelly, and A.~J.
  Davison, ``{SLAM}++: Simultaneous localisation and mapping at the level of
  objects,'' \emph{2013 IEEE Conference on Computer Vision and Pattern
  Recognition}, 2013.

\bibitem{grinvald2019volumetric}
M.~{Grinvald}, F.~{Furrer}, T.~{Novkovic}, J.~J. {Chung}, C.~{Cadena},
  R.~{Siegwart}, and J.~{Nieto}, ``{Volumetric Instance-Aware Semantic Mapping
  and 3D Object Discovery},'' \emph{IEEE Robotics and Automation Letters},
  2019.

\bibitem{Rosinol20icra-Kimera}
A.~Rosinol, M.~Abate, Y.~Chang, and L.~Carlone, ``Kimera: an open-source
  library for real-time metric-semantic localization and mapping,'' in
  \emph{2020 IEEE International Conference on Robotics and Automation (ICRA)},
  2020.

\bibitem{hughes2022hydra}
N.~Hughes, Y.~Chang, and L.~Carlone, ``Hydra: A real-time spatial perception
  system for {3D} scene graph construction and optimization,'' in \emph{2022
  Robotics: Science and Systems (RSS)}, 2022.

\bibitem{fusion++}
J.~McCormac, R.~Clark, M.~Bloesch, A.~Davison, and S.~Leutenegger, ``Fusion++:
  Volumetric object-level {SLAM},'' \emph{2018 International Conference on 3D
  Vision (3DV)}, 2018.

\bibitem{SemanticFusionD3}
J.~McCormac, A.~Handa, A.~Davison, and S.~Leutenegger, ``Semanticfusion: Dense
  3d semantic mapping with convolutional neural networks,'' \emph{2017 IEEE
  International Conference on Robotics and Automation (ICRA)}, 2017.

\bibitem{Rnz2018MaskFusionRR}
M.~R{\"u}nz and L.~de~Agapito, ``Maskfusion: Real-time recognition, tracking
  and reconstruction of multiple moving objects,'' \emph{2018 IEEE
  International Symposium on Mixed and Augmented Reality (ISMAR)}, 2018.

\bibitem{Sun2019MovableObjectAwareVS}
T.~Sun, Y.~Sun, M.~Liu, and D.-Y. Yeung, ``Movable-object-aware visual {SLAM}
  via weakly supervised semantic segmentation,'' \emph{ArXiv}, 2019.

\bibitem{dsg}
A.~Rosinol, M.~Abate, J.~Shi, and L.~Carlone, ``3d dynamic scene graphs:
  Actionable spatial perception with places, objects, and humans,'' in
  \emph{Robotics: Science and Systems}, 2020.

\bibitem{DM-SLAM}
X.~Lu, H.~Wang, S.~Tang, H.~Huang, and C.~Li, ``Dm-slam: Monocular {SLAM} in
  dynamic environments,'' \emph{Applied Sciences}, 2020.

\bibitem{DOT}
I.~Ballester, A.~Font{\'{a}}n, J.~Civera, K.~H. Strobl, and R.~Triebel,
  ``{DOT:} dynamic object tracking for visual {SLAM},'' \emph{CoRR}, 2020.

\bibitem{DS_SLAM}
C.~Yu, Z.~Liu, X.~Liu, F.~Xie, Y.~Yang, Q.~Wei, and F.~Qiao, ``{DS-SLAM:} {A}
  semantic visual {SLAM} towards dynamic environments,'' \emph{CoRR}, 2018.

\bibitem{Hachiuma2019DetectFusionDA}
R.~Hachiuma, C.~Pirchheim, D.~Schmalstieg, and H.~Saito, ``Detectfusion:
  Detecting and segmenting both known and unknown dynamic objects in real-time
  {SLAM},'' in \emph{BMVC}, 2019.

\bibitem{TSDF++}
M.~Grinvald, F.~Tombari, R.~Y. Siegwart, and J.~I. Nieto, ``{TSDF++}: A
  multi-object formulation for dynamic object tracking and reconstruction,''
  \emph{2021 IEEE International Conference on Robotics and Automation (ICRA)},
  2021.

\bibitem{Xu2019MIDFusionOO}
B.~Xu, W.~Li, D.~Tzoumanikas, M.~Bloesch, A.~J. Davison, and S.~Leutenegger,
  ``{MID}-fusion: Octree-based object-level multi-instance dynamic {SLAM},''
  \emph{2019 International Conference on Robotics and Automation (ICRA)}, 2019.

\bibitem{visinsMulti}
Y.~Ren, B.~Xu, C.~L. Choi, and S.~Leutenegger, ``Visual-inertial multi-instance
  dynamic {SLAM} with object-level relocalisation,'' in \emph{arXiv}, 2022.

\bibitem{Fu2022RobustCD}
\BIBentryALTinterwordspacing
J.~Fu, Y.~Du, K.~Singh, J.~B. Tenenbaum, and J.~J. Leonard, ``Robust change
  detection based on neural descriptor fields,'' \emph{Proc. of the IEEE/RSJ
  International Conference on Intelligent Robots and Systems (IROS)}, pp.
  2817--2824, 2022. [Online]. Available:
  \url{https://api.semanticscholar.org/CorpusID:251223475}
\BIBentrySTDinterwordspacing

\bibitem{Looper20223DVL}
\BIBentryALTinterwordspacing
S.~Looper, J.~R. Puigvert, R.~Y. Siegwart, C.~Cadena, and L.~M. Schmid, ``3d
  vsg: Long-term semantic scene change prediction through 3d variable scene
  graphs,'' \emph{Proc. of the IEEE International Conference on Robotics and
  Automation (ICRA)}, pp. 8179--8186, 2022. [Online]. Available:
  \url{https://api.semanticscholar.org/CorpusID:252355114}
\BIBentrySTDinterwordspacing

\bibitem{rosen2016towards}
D.~M. Rosen, J.~Mason, and J.~J. Leonard, ``Towards lifelong feature-based
  mapping in semi-static environments,'' in \emph{2016 IEEE International
  Conference on Robotics and Automation (ICRA)}.\hskip 1em plus 0.5em minus
  0.4em\relax IEEE, 2016.

\bibitem{berkenkamp2015safe}
F.~Berkenkamp and A.~P. Schoellig, ``Safe and robust learning control with
  gaussian processes,'' in \emph{Proc. of the European Control Conference
  (ECC)}, 2015, pp. 2496--2501.

\bibitem{joshi2019deep}
G.~Joshi and G.~Chowdhary, ``Deep model reference adaptive control,'' in
  \emph{Proc. of the IEEE Conference on Decision and Control (CDC)}, 2019, pp.
  4601--4608.

\bibitem{taylor2020learning}
A.~Taylor, A.~Singletary, Y.~Yue, and A.~Ames, ``Learning for safety-critical
  control with control barrier functions,'' in \emph{Proc. of the Learning for
  Dynamics and Control Conference}, 2020, pp. 708--717.

\bibitem{richards2018lyapunov}
S.~M. Richards, F.~Berkenkamp, and A.~Krause, ``The {L}yapunov neural network:
  Adaptive stability certification for safe learning of dynamical systems,'' in
  \emph{Proc. of the Conference on Robot Learning}, 2018, pp. 466--476.

\bibitem{dean2021guaranteeing}
S.~Dean, A.~Taylor, R.~Cosner, B.~Recht, and A.~Ames, ``Guaranteeing safety of
  learned perception modules via measurement-robust control barrier
  functions,'' in \emph{Proc. of the Conference on Robot Learning}, 2021, pp.
  654--670.

\bibitem{Jian2022DynamicCB}
\BIBentryALTinterwordspacing
Z.~Jian, Z.~Yan, X.~Lei, Z.-R. Lu, B.~Lan, X.~Wang, and B.~Liang, ``Dynamic
  control barrier function-based model predictive control to safety-critical
  obstacle-avoidance of mobile robot,'' \emph{2023 IEEE International
  Conference on Robotics and Automation (ICRA)}, pp. 3679--3685, 2022.
  [Online]. Available: \url{https://api.semanticscholar.org/CorpusID:252367490}
\BIBentrySTDinterwordspacing

\bibitem{zeng2021safety}
J.~Zeng, B.~Zhang, and K.~Sreenath, ``Safety-critical model predictive control
  with discrete-time control barrier function,'' in \emph{Proc. of the American
  Control Conference (ACC)}, 2021, pp. 3882--3889.

\bibitem{Singletary2020ComparativeAO}
\BIBentryALTinterwordspacing
A.~W. Singletary, K.~Klingebiel, J.~R. Bourne, A.~W. Browning, P.~T. Tokumaru,
  and A.~D. Ames, ``Comparative analysis of control barrier functions and
  artificial potential fields for obstacle avoidance,'' \emph{2021 IEEE/RSJ
  International Conference on Intelligent Robots and Systems (IROS)}, pp.
  8129--8136, 2020. [Online]. Available:
  \url{https://api.semanticscholar.org/CorpusID:224803757}
\BIBentrySTDinterwordspacing

\bibitem{Han2019FIESTAFI}
\BIBentryALTinterwordspacing
L.~Han, F.~Gao, B.~Zhou, and S.~Shen, ``Fiesta: Fast incremental euclidean
  distance fields for online motion planning of aerial robots,'' \emph{2019
  IEEE/RSJ International Conference on Intelligent Robots and Systems (IROS)},
  pp. 4423--4430, 2019. [Online]. Available:
  \url{https://api.semanticscholar.org/CorpusID:70350033}
\BIBentrySTDinterwordspacing

\bibitem{pybullet}
E.~Coumans and Y.~Bai, ``Pybullet, a python module for physics simulation for
  games, robotics and machine learning,'' \url{http://pybullet.org},
  2016--2021.

\bibitem{Karaman2011AnytimeMP}
\BIBentryALTinterwordspacing
S.~Karaman, M.~R. Walter, A.~Perez, E.~Frazzoli, and S.~J. Teller, ``Anytime
  motion planning using the rrt*,'' \emph{2011 IEEE International Conference on
  Robotics and Automation}, pp. 1478--1483, 2011. [Online]. Available:
  \url{https://api.semanticscholar.org/CorpusID:1003020}
\BIBentrySTDinterwordspacing

\end{thebibliography}
